%% file: arxiv_main.tex
\documentclass[lettersize,journal]{IEEEtran}
\pdfoutput=1 
\usepackage{array}
\usepackage{amsmath,amsfonts}
\usepackage{algorithmic}
\usepackage{algorithm}
\usepackage[caption=false,font=normalsize,labelfont=sf,textfont=sf]{subfig}
\usepackage{textcomp}
\usepackage{stfloats}
\usepackage{url}
\usepackage{verbatim}
\usepackage{graphicx}
\usepackage{caption}
\usepackage{cite}
\usepackage{amssymb}
\usepackage{amsthm}
\usepackage{latexsym}
\usepackage{url}

\usepackage[breaklinks=true,colorlinks,citecolor=blue,urlcolor=blue,linkcolor=blue]{hyperref}

\usepackage{preamble/beimath}
\usepackage{preamble/textoolz}

\usepackage{comment}
\usepackage{xcolor}
\usepackage{orcidlink}

\usepackage{booktabs}
\usepackage{multirow}
\usepackage{makecell}
\usepackage{color, colortbl}
\definecolor{Gray}{gray}{0.9}
\usepackage[capitalize]{cleveref}

\newcommand{\et}[2]{${#1}^{\pm{#2}}$}
\newcommand{\etb}[2]{$\mathbf{{#1}}^{\pm{#2}}$}

\crefname{table}{Tab.}{Tabs}
\Crefname{table}{Tab}{Tabs}
\crefname{figure}{Fig.}{Figs.}
\Crefname{figure}{Fig.}{Figs.}
\crefname{equation}{Eq.}{Eqs.}
\Crefname{equation}{Eq.}{Eqs.}

\newif\ifshowchanges
\showchangesfalse   

\ifshowchanges
  \newcommand{\re}[1]{\textcolor{blue}{#1}}  
\else
  \newcommand{\re}[1]{#1}                    
\fi

\newif\ifshowchangesTwo
\showchangesTwofalse   

\ifshowchangesTwo
  \newcommand{\reTwo}[1]{\textcolor{blue}{#1}}  
\else
  \newcommand{\reTwo}[1]{#1}                    
\fi

\begin{document}

\title{MARRS: Masked Autoregressive Unit-based Reaction Synthesis}
\author{Yabiao Wang\textsuperscript{*}, Shuo Wang\textsuperscript{*}, Jiangning Zhang, Jiafu Wu, Qingdong He, Yong Liu\textsuperscript{$\dagger$}

\thanks{Digital Object Identifier 10.1109/TVCG.2026.3675978}
\thanks{Yabiao Wang is with the Institute of Cyber-Systems and Control, Zhejiang University, Hangzhou, 310027, China, and Tencent Youtu Lab, Shanghai, 200030, China (e-mail: yabiaowang@zju.edu.cn).} 
\thanks{Shuo Wang, Jiangning Zhang, Jiafu Wu, Qingdong He are with the Tencent Youtu Lab, Shanghai, 200030, China (e-mail: \{leifwang, vtzhang, jiafwu, yingcaihe\}@tencent.com).}
\thanks{Yong Liu is with the Institute of Cyber-Systems and Control, Zhejiang University, Hangzhou, 310027, China (e-mail: yongliu@iipc.zju.edu.cn).} 
\thanks{$^*$Equal contribution.~~~~~$^\dagger$Corresponding author.}
}

\markboth{IEEE TRANSACTIONS ON VISUALIZATION AND COMPUTER GRAPHICS}%
{Shell \MakeLowercase{\textit{et al.}}: A Sample Article Using IEEEtran.cls for IEEE Journals}

\maketitle

\input{sec/0_abstract}    


\input{sec/1_intro}
\input{sec/2_relatedwork}
\input{sec/3_method}

\input{sec/4_experiments}

\bibliographystyle{IEEEtran}
\bibliography{main.bib}


\end{document}

%% file: sec/0_abstract.tex
\begin{abstract}
This work aims at a challenging task: human action-reaction synthesis, \emph{i.e.}, generating human reactions conditioned on the action sequence of another person. Currently, autoregressive modeling approaches with vector quantization (VQ) have achieved remarkable performance in motion generation tasks. However, VQ has inherent disadvantages, including quantization information loss, low codebook utilization, \emph{etc}. 
In addition, while dividing the body into separate units can be beneficial, the computational complexity needs to be considered. Also, the importance of mutual perception among units is often neglected.
In this work, we propose MARRS, a novel framework designed to generate coordinated and fine-grained reaction motions using continuous representations. 
Initially, we present the Unit-distinguished Motion Variational AutoEncoder (UD-VAE), which segments the entire body into distinct body and hand units, encoding each independently. Subsequently, we propose Action-Conditioned Fusion (ACF), which involves randomly masking a subset of reactive tokens and extracting specific information about the body and hands from the active tokens. Furthermore, \re{we introduce Mutual Unit Modulation (MUM) to facilitate interaction between body and hand units by using the information from one unit to adaptively modulate the other.} Finally, for the diffusion model, we employ a compact MLP as a noise predictor for each distinct body unit and incorporate the diffusion loss to model the probability distribution of each token.
Both quantitative and qualitative results demonstrate that our method achieves superior performance. Project page: \textcolor{magenta}{\url{https://aigc-explorer.github.io/MARRS/}}.
\end{abstract}

%% file: sec/1_intro.tex
\section{Introduction}
\label{sec:intro}

\IEEEPARstart{I}n the field of generative computer vision, \re{human-human motion generation holds substantial importance for computer animation~\cite{parent2012computer, Animation1_TVCG, magnenat1985computer}, game development~\cite{urbain2010introduction,MOST_TVCG, bethke2003game}, and robotic control~\cite{saridis1983intelligent,wang2023towards, chen2023learning, wang2024sqd}. 
In this work, we focus on generative models for human action-reaction synthesis, \emph{i.e.}, generating human reactions based on the action sequence of another individual as conditions}, which can significantly reduce the workload for animators by enabling them to design an acting character and automatically generate meaningful motions for a reacting character.

Although autoregressive modeling approaches have recently achieved remarkable performance in motion generation tasks~\cite{pinyoanuntapong2024mmm}, it is sub-optimal to apply these methods to human action-reaction synthesis. \re{For example, our experiments demonstrate that the reconstruction and generation performance of VQ-VAE are unsatisfactory.}
There are two challenges for this:
first, vector quantization (VQ) exhibits inherent limitations. Mapping continuous motion data to a constrained set of discrete tokens inevitably results in a loss of information. Although Momask~\cite{guo2024momask} utilizes successive stages of residual quantization to minimize quantization errors incrementally, complex multistage training does not fundamentally eliminate quantization errors. Second, the notorious codebook collapse~\cite{zheng2023cvq, EdVAE_PR} makes optimizing the code vectors in the existing VQ-VAE not entirely trivial. 
Without relying on VQ-VAE, some methods~\cite{shi2024amdm} have tried autoregressive diffusion in motion generation, but their cumbersome training and inference processes do not achieve stunning performance in single-segment motion generation.
%
Moreover, units splitting has been shown to be an effective strategy in motion generation. However, as the number of units increases, the computational effort of the model also increases. And the importance of mutual perception among units is often neglected. Thus, it is necessary to explore a more rational units splitting as well as to design an efficient and effective units perception method.

\begin{figure*}[t]
  \centering
   \includegraphics[width=1.0\linewidth]{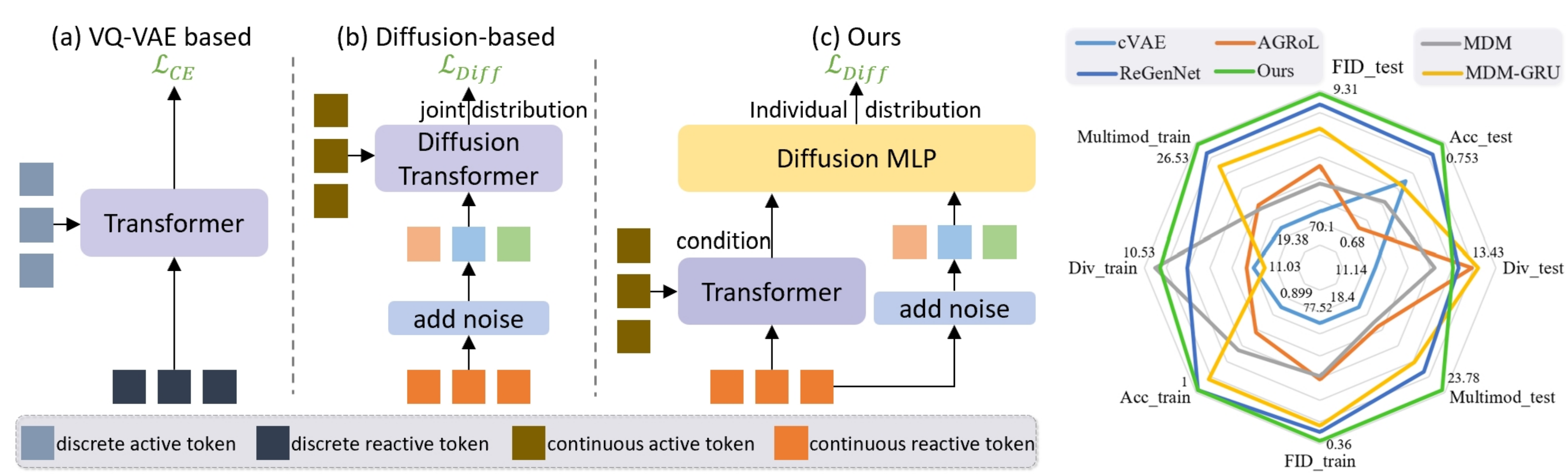}
   \caption{
   \textbf{Left: Paradigm comparison of different frameworks.}
   (a) and (b) present the structures of the VQ-VAE-based and Diffusion-based methods, respectively, while (c) shows the framework of our proposed MARRS. $\mathcal{L}_{CE}$ is cross entropy loss, $\mathcal{L}_{Diff}$ is diffusion loss.
   \textbf{Right: result comparison among our method and other methods on eight metrics.}
}
   \label{fig:teaser}
\end{figure*}

\re{Recently, some works~\cite{li2024mar} have explored autoregressive image generation without vector quantization, providing new insight into autoregressive generation methods. In this paper, we propose MARRS, a novel framework that generates synchronized and fine-grained reaction movements in continuous representations. The framework has two advantages. First, compared with VQ-based frameworks, our motion representation operates in a continuous space, which helps to avoid the quantization errors associated with vector quantization(VQ). Second, diffusion-based frameworks typically model the joint distribution of all tokens within a single denoising process, whereas our framework factorizes the problem and models the conditional distribution of one token at a time. This decomposition empirically eases optimization and is more conducive to effective learning. As shown in~\cref{fig:teaser}, we provide a paradigm comparison of different frameworks and a result comparison among different methods.}

During the training phase, MARRS follows a two-stage paradigm. In the first stage, for the network to learn the concept of the body and hands, we propose Unit-distinguished Motion Variational AutoEncoder (UD-VAE), which splits the whole-body into body and hands units and then encodes them independently using a variational autoencoder. In the second stage, we propose Action-Conditioned Fusion (ACF), which involves randomly masking out a portion of the reactive tokens and then extracting the information of body and hands from the active tokens separately using the transformer. Next, in order to avoid the lack of mutual perception between the two units (body and hands), we propose Mutual Unit Modulation (MUM), which utilizes information from the body or hands to modulate the other unit adaptively and enables interaction between the body and hands generators to jointly produce whole-body motion. Finally, we utilize a compact MLP to serve as a noise predictor for each distinct unit, modeling individual tokens separately and integrating the diffusion loss~\cite{li2024mar} to assess the probability of each token.

Our contributions can be summarized as follows:

\begin{itemize}
\item we propose MARRS, a novel framework that generates synchronized and fine-grained reaction movements. To the best of our knowledge, this is the first successful application of masked autoregressive generation to the field of action-reaction synthesis.

\item We introduce the Unit-distinguished Motion Variational AutoEncoder (UD-VAE) to enable the network to grasp the concepts of the body and hands. Next, we propose Action-Conditioned Fusion (ACF), which could extract information about the body and hands of the reactor from the active tokens. Furthermore, to prevent the body and hand units from lacking awareness of each other, we propose Mutual Unit Modulation (MUM). Finally, we employ a compact MLP as a noise predictor and integrate the diffusion loss to estimate the probability of each token.


\item We conduct extensive experiments on the human action-reaction synthesis datasets, NTU120-AS and Chi3D-AS. The results in both online and offline settings demonstrate the effectiveness of our proposed methods.
\end{itemize}

%% file: sec/2_relatedwork.tex
\section{Related Work}
\label{sec:formatting}

\subsection{Motion Generation Framework}

TEMOS~\cite{petrovich2022temos} and T2M~\cite{guo2022t2m} use a Transformer-based VAE combined with a text encoder to create motion sequences from textual descriptions. T2M-GPT~\cite{zhang2023t2mgpt} introduces a framework based on VQ-VAE and Generative Pretrained Transformer (GPT) for motion generation. MMM~\cite{pinyoanuntapong2024mmm} proposes a conditional masked motion architecture based on VQ-VAE. To mitigate quantization errors, MoMask~\cite{guo2024momask} introduces a hierarchical quantization scheme.
MotionDiffuse~\cite{zhang2024motiondiffuse} and MDM~\cite{tevet2209mdm} are groundbreaking frameworks for text-driven motion generation using diffusion models. 
Instead of using a diffusion model to connect raw motion sequences with conditional inputs, MLD~\cite{chen2023mld} employs a latent diffusion model to significantly reduce training and inference costs. ReMoDiffuse~\cite{zhang2023remodiffuse} introduces an enhancement mechanism based on dataset retrieval to improve the denoising process in diffusion models. SPORT~\cite{SPOR_TVCG} integrates specialized encoders and a diffusion-based decoder to generate real-time, diverse motions from zero-shot text prompts. GUESS~\cite{GUESS_TVCG} introduces a novel cascaded diffusion framework for text-driven human motion synthesis. It employs a multi-stage generation process that recursively abstracts human poses into coarser skeletons, significantly improving motion stability and detail. 

\subsection{Human-Human Motion Generation}

Synthesizing human-human interactions is essential for applications in gaming and augmented/virtual reality. Earlier approaches often utilized motion graphs~\cite{shum2007simulating} and momentum-based inverse kinematics~\cite{komura2005animating} to simulate the movements of human joints. Recently, more sophisticated techniques have emerged. ComMDM~\cite{shafir2023commdm} refines a pre-trained text-to-motion diffusion model using a limited dataset of two-person motions. RIG~\cite{tanaka2023rig} transforms the text describing asymmetric interactions into both active and passive forms to ensure consistent textual context for each participant. To facilitate the interaction by positioning the joints of two individuals accurately, InterControl~\cite{wang2023intercontrol} leverages a Large Language Model to create movement plans through carefully crafted prompts. InterGen~\cite{liang2024intergen} introduces a diffusion model with shared parameters and several regularization losses. FreeMotion~\cite{fan2024freemotion} proposes to integrate the single and multi-person motion by the conditional motion distribution. Designing an efficient approach to modeling between two people, TIMotion~\cite{wang2024timotion} proposes a temporal and interactive framework. \re{InterMask~\cite{javed2025intermask} proposes a novel generative framework that uses collaborative masked modeling in a discrete 2D motion token space, supporting reaction generation without additional fine-tuning. Based on the premise of actor's motion, ReGenNet~\cite{xu2024regennet} proposes a diffusion-based generative model and explicit distance-based interaction loss. Reactffusion~\cite{zhang2025reactffusion} improves reaction motion generation by using predicted physical contact maps to guide a diffusion model, reducing joint mismatches and providing a plug-and-play enhancement for other methods.} 

\subsection{Autoregressive Modeling without Quantization}
MAR~\cite{li2024mar} initially utilizes autoregressive models within a continuous-valued domain to represent the probability distribution of each token through a diffusion process. MarDini~\cite{liu2024mardini} employs MAR for temporal planning, allowing for video generation based on any number of masked frames at various frame positions. 
MMAR~\cite{yang2024mmar} processes continuous-valued image tokens to prevent information loss and separates the diffusion process from the autoregressive model in the context of multi-modal large language models. AMDM~\cite{shi2024amdm} utilizes the diffusion model to implement frame-by-frame incremental generation. MARDM~\cite{meng2024mardm} presents a human motion diffusion model capable of executing bidirectional masked autoregression. DPS~\cite{in_betw_TVCG} employs conditional variational autoencoders (CVAEs) to implement their autoregressive networks to generate high-quality
and diverse transitions simultaneously for in-betweening task.

%% file: sec/3_method.tex
\section{Method}

\subsection{Problem Formulation}

\begin{figure*}[h]
  \centering
   \includegraphics[width=1.0\linewidth]{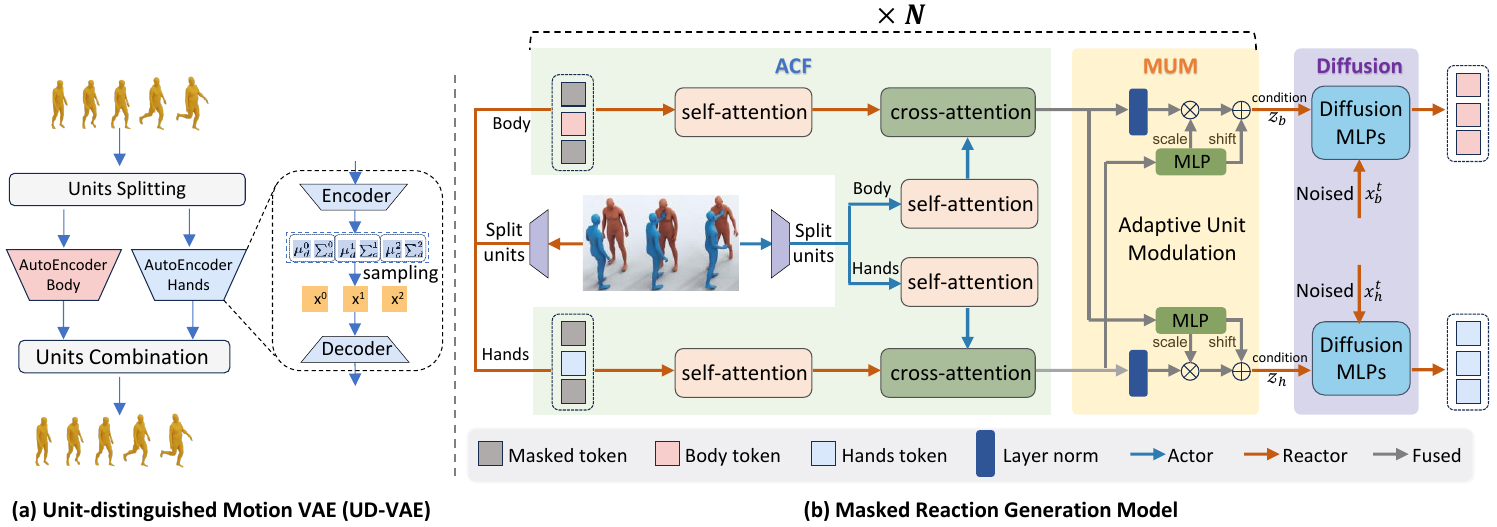}
   \caption{\re{ \textbf{The overall framework of our proposed MARRS.} (a) Whole-body motion is divided into two units: body and hands and then each unit is encoded independently by a VAE. (b) shows the process of the masked reaction generation model. First, the reactive token of each unit obtains the interaction information from the active token through Action-Conditioned Fusion (ACF). Then different units acquire the coordinated information through Mutual Unit Modulation (MUM). ACF and MUM form the basic model blocks, of which there are N (8 in MARRS-Base). Finally, we employ a compact MLP in the Diffusion model to serve as a noise predictor and integrate the diffusion loss to model the probability of each token.}
}
   \label{fig:network}
\end{figure*}

In the context of synthesizing human action-reaction sequences, our objective is to produce a reaction that is conditioned on a given action. \re{Formally, given that the reaction sequence $x^{1:N} = \{x^i\}_{i=1}^N$ and the action sequence $y^{1:N} = \{y^i\}_{i=1}^N$, where N is sequence length.} Our goal is to model the conditional probability distribution ${\rm P}(x^{1:N} | y^{1:N})$, enabling us to sample realistic reactive motions. As with~\cite{xu2024regennet}, we utilize the SMPL-X~\cite{SMPL-X-2019} human model to represent the sequence of human motions. Thus, the reaction can be represented as $x^i = [\theta^x_i, q^x_i, \gamma^x_i]$, where $\theta^x_i \in \mathbb{R}^{3K}$, $q^x_i \in \mathbb{R}^{3} $, $\gamma^x_i \in \mathbb{R}^{3} $ are the pose parameters, global orientation, and root translation of the person. Here, $K$ is set to 54, which represents the number of body joints, including the jaw, eyeballs, and fingers.

\subsection{Overview of MARRS}
Our method consists of two stages to generate the reactive motions on the premise that one's actions are known. In the first stage, we split the whole-body motion into two units: body and hands, and then map them to continuous-valued tokenizers. In the second stage, the reactive token of each unit initially gathers information from the active token via Action-Conditioned Fusion. Subsequently, different units gain coordinated information through Mutual Unit Modulation. Finally, we utilize a compact MLP as a noise predictor and incorporate the diffusion loss to model each token. The overall structure is shown in ~\cref{fig:network}.

\subsection{Unit-distinguished Motion VAE (UD-VAE)}
\label{method_vae}
In the first stage, we split the whole-body motion into two units: body and hands. Then each unit is encoded independently by a variational autoencoder (VAE), thereby providing prior knowledge about the concept of body and hands.

Specifically, given an active motion sequence $y^{1:N}$, we separate it into two units $\{y^{1:N}_k, k\in[body, hands] \}$. Then the unit $y^{1:N}_k$ is fed into the corresponding $Encoder_k$ to obtain the embedding $E^{1:L}_k$, where $L=\frac{N}{r}$ and $r$ is the downsampling rate of the encoder. Next, the Gaussian distribution parameters $\mu_k^{1:L}$ and $\sum_k^{1:L}$ are obtained through a linear layer, and we sample latent vectors $z \in \mathbb{R}^d$ from the above distributions. The latent dimension d is set to 256 in our experiments. Finally, the sampled vectors are input into the $Decoder_k$ to obtain the reconstructed $k$-th unit $\hat{y}^{1:N}_k$. 

Similarly, we can acquire the reconstructed $k$-th unit $\hat{x}^{1:N}_k$ for the reactive motion sequence $x^{1:N}$. The
optimization objective of the $k$-th unit’s VAE is: 
\begin{equation}
\resizebox{0.90\linewidth}{!}{$
\begin{aligned}
\mathcal{L}_{VAE}^k & = {\rm SmoothL1} (\hat{x}^{1:N}_k,x^{1:N}_k) + 
{\rm SmoothL1}(\hat{y}^{1:N}_k,y^{1:N}_k). 
\label{equ:noise_loss}
\end{aligned}
$}
\end{equation}
\re{The structure of the UD-VAE is shown in the overall framework. Both the encoder and the decoder use 1D ResNet backbones, following the design of T2M-GPT~\cite{zhang2023t2mgpt}.}

\subsection{Masked Reaction Generation Model}
\re{In the second stage, we introduce the reaction generation framework as shown in the overall structure, which consists of Action-Conditioned Fusion, Mutual Unit Modulation, and diffusion MLPs belonging to their respective units.} 

\subsubsection{\textbf{Action-Conditioned Fusion (ACF)}}
\label{method1}
During training, action and reaction are sent to UD-VAE respectively, and continuous-valued tokenizers corresponding to body and hands are obtained after the encoder. 
To enable the networks to be aware of the conception of different whole-body units, we use two networks with the same structure but no shared parameters for the body and hands. First, we define the computation of $\rm Attn$ as:
\begin{equation}
\resizebox{0.9\linewidth}{!}{$
\begin{aligned}
  {\rm Attn}({\rm Q, K, V}) = softmax(\frac{({\rm Q W^Q}) \cdot ({\rm K W^K})^T}{\sqrt{C}}) \cdot ({\rm V W^V}),
\end{aligned}
$}
\end{equation}
where ${\rm W^Q}$, ${\rm W^K}$, and ${\rm W^V}$ are trainable weights.

Specifically, given the continuous-valued \textit{\textbf{body tokens}} of actor $Y_{b}$ and reactor $X_{b}$, we further extract the actor's motion information $Y_{b}$ through $\rm Attn$ and obtain the refined token embeddings $Y_{b}^{\prime}$ as: 
\begin{align}
  Y_{b}^{\prime} = {\rm Attn}(Y_{b}, Y_{b}, Y_{b}).
\end{align}
Next, we randomly mask out a varying fraction of sequence elements $X_{b}$ by replacing the tokens with a special [MASK] token and the resulting sequence is denoted as $\hat{X}_{b}$. Our goal is to predict the masked tokens given $\hat{X}_{b}$ and the active tokens $Y_{b}^{\prime}$. 
Then the reactive token embedding $\hat{X}_{b}^{fusion}$ incorporating the action information can be obtained as follows:

\begin{align}
  \hat{X}_{b}^{\prime} &= {\rm Attn}(\hat{X}_{b}, \hat{X}_{b}, \hat{X}_{b}),\\
  \hat{X}_{b}^{fusion} &= {\rm Attn}(\hat{X}_{b}^{\prime}, Y_{b}^{\prime}, Y_{b}^{\prime}).
\end{align}

For the continuous-valued \textit{\textbf{hands tokens}} ($Y_{h}$, $X_{h}$) of the actor and reactor, we can acquire the reactive token embedding $\hat{X}_{h}^{fusion}$ in the same way as above.

\subsubsection{\textbf{Mutual Unit Modulation (MUM)}} 
 The whole-body motion can be divided into two units: the body unit and the hands unit. As noted in~\cite{zou2024parco}, depending exclusively on these individual unit motion generators may fail to produce coordinated whole-body motions, as they lack awareness of other units. Hence, we propose MUM, which enables interaction between the body and hands generators to jointly produce whole-body motion.

Given that the reactive token embeddings $\hat{X}_{b}^{fusion}$ and $\hat{X}_{h}^{fusion}$ from ACF, \re{we utilize the body's information to modulate and optimize the position of the hands. The modulated hands embeddings $\hat{X}_{h}^{final}$ are acquired as follows:}
\begin{align}
  scale_b, shift_b = {\rm Linear}(\hat{X}_{b}^{fusion}),\\ 
  \hat{X}_{h}^{final} = scale_b \cdot {\rm LN}(\hat{X}_{h}^{fusion}) + shift_b.
\end{align}
where $\rm LN$ denotes the layer normalization.
Similarly, we can obtain the modulated body embeddings $\hat{X}_{b}^{final}$ as:
\begin{align}
  scale_h, shift_h = {\rm Linear}(\hat{X}_{h}^{fusion}),\\ 
  \hat{X}_{b}^{final} = scale_h \cdot {\rm LN}(\hat{X}_{b}^{fusion}) + shift_h.
\end{align}

\re{ACF and MUM form the basic model blocks, of which there are N (8 in MARRS-Base)}.

\subsubsection{\textbf{\re{Processing in Diffusion MLP}}} 
\re{Taking the body branch of MUM as an example, the output of MUM $z_b$ serves as a condition.
\begin{align}
U = z_b + t,
\end{align}
where t is the time step in diffusion process.
Three linear layers with the same structure but different parameters are adopted.
\begin{gather}
U_{\lambda} = {\rm Linear}_{\lambda}(U),\\
U_{\gamma}  = {\rm Linear}_{\gamma}(U), \\
U_{\beta}   = {\rm Linear}_{\beta}(U).
\end{gather}
We employ a 3-layer MLP as the noise estimator $\epsilon_{\theta_1}$, which is formulated as:
\begin{equation}
{\epsilon}_{\theta_1}\left({x}_{b}^{t} \mid t, {z}_{b}\right)=\mathbf{MLP}(\cdot ).
\end{equation}
During training, ${x}_{b}^{t}$ is the noise-corrupted body motion; during inference, ${x}_{b}^{t}$ is initialized as Gaussian noise $\mathcal{N}(\mathbf{0}, \mathbf{I})$.
$t$ and ${z}_{b}$ are used as conditions, and the calculation process of $\mathbf{MLP}(\cdot )$ is as follows:
\begin{gather}
U_1 = \Phi(x_b^{t}) = x_b^{t}+U_{\lambda} f \left( U_{\gamma}{\rm LN}\left( x_b^{t} \right)+ U_{\beta}\right),\\
U_2=\Phi(U_1),\\ 
\mathbf{MLP}(\cdot )=U_3=\Phi(U_2).
\end{gather}
Where $f$ is a two-layer linear projection with non-linear activation, and LN denotes the layer normalization. $\Phi(\cdot)$ can be considered a basic block in this process.
The hands branch in MUM follow the same structure but with different parameters.}

\subsubsection{\textbf{Diffusion for Autoregressive Reaction Generation}} 
Autoregressive methods~\cite{pinyoanuntapong2024mmm, guo2024momask} have shown considerable benefits in motion generation. Instead of modeling the jointed distribution of all tokens, autoregressive models formulate the generation task as ``next token prediction”:
\begin{align}
  p(x^1, _{\cdots}, x^n ) = \prod _{i=1}^n p(x^i | x^1, _{\cdots}, x^{i-1} ),
\end{align}
where $\{x^1, x^2, _{\cdots}, x^n\}$ is the sequence of tokens and the superscript $1 \leq i \leq n$ specifies an order. 

As noted in MAR~\cite{li2024mar}, training diffusion models by employing MSE loss to carry out autoregressive generation leads to a disastrous FID score because of the failure of capturing chained probability distributions. Motivated by the successful application of MAR in image generation, we incorporate the diffusion loss~\cite{li2024mar} to estimate the probability of each token within the field of the reaction generation.

Given that the modulated body embedding $\hat{X}_{b} \in \mathbb{R}^{L \times d}$ and hands embedding $\hat{X}_{h} \in \mathbb{R}^{L \times d}$, where $L$ is the number of tokens. Each token we sample from $\hat{X}_{b}$ and $\hat{X}_{h}$ are denoted as $z_{b} \in \mathbb{R}^{d}$ and $z_{h} \in \mathbb{R}^{d}$, and we define the corresponding ground truth tokens as $x_{b}$ and $x_{h}$. The loss function can be formulated as a denoising criterion:
\begin{align}
\label{eq:diffusion_loss}
\mathcal{L}({x}\,|\,{z})=
\mathbb{E}_{\varepsilon_1, t}\left[\left\|\epsilon_1-{\epsilon}_{\theta_1}\left({x}_{b}^{t} \mid t, {z}_{b}\right)\right\|^{2}\right]\, + \nonumber \\
\mathbb{E}_{\varepsilon_2, t}\left[\left\|\epsilon_2-{\epsilon}_{\theta_2}\left({x}_{h}^{t} \mid t, {z}_{h}\right)\right\|^{2}\right]\, .
\end{align}


Here $\epsilon_1 \in \mathbb{R}^{d}$ and $\epsilon_2 \in \mathbb{R}^{d}$ are Gaussian vectors sampled from $\mathcal{N}(\mathbf{0}, \mathbf{I})$. The noise-corrupted data are ${x}_{b}^{t} = \sqrt{\bar{\alpha}_{t}} {x_{b}}+\sqrt{1-\bar{\alpha}_{t}} \epsilon_1$ and ${x}_{h}^{t} = \sqrt{\bar{\alpha}_{t}} {x_{h}}+\sqrt{1-\bar{\alpha}_{t}} \epsilon_2$, where $\bar{\alpha}_{t}$ is a noise schedule~\cite{nichol2021cosnoise} indexed by a time step $t$. We model individual tokens by using a small MLP as a noise estimator for each unit separately. The noise estimators $\epsilon_{\theta_1}$ and $\epsilon_{\theta_2}$ have the same structure but do not share parameters. The notation $\epsilon_{\theta}({x}_{n}^{t}\, |\, t, {z}_{n})$ means that this network takes ${x}_{n}^{t}$ as the input, and is conditional on both ${t}$ and ${z}_{n}$.

\subsubsection{\textbf{Inference}} 

\begin{figure*}[htbp]
  \centering
   \includegraphics[width=1.0\linewidth]{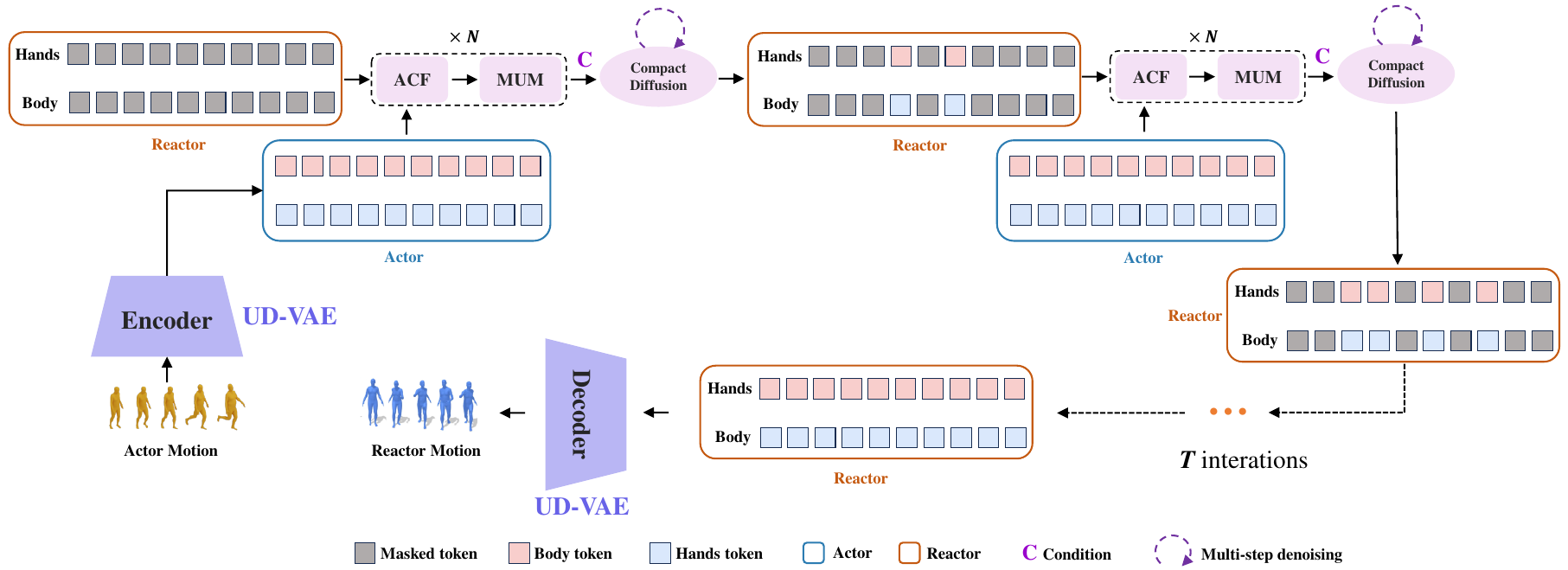}
      \caption{ \re{\textbf{Visualization of inference process.} The generation of entire tokens is performed in an autoregressive manner. \textbf{Compact diffusion} model is very small, consisting of only a 3-layer MLP. Therefore, MARRS can achieve fast inference speed.}
      }
   \label{fig:infer_process}
\end{figure*}

\re{The inference pipeline is illustrated in \cref{fig:infer_process}. Given an active motion sequence, we encode it with UD-VAE to obtain active tokens, then autoregressively generate reactive tokens over $T$ iterations, where the mask ratio at iteration $t$ follows a cosine decay schedule, $\cos\!\left(\frac{\pi t}{2T}\right)$.}

First, we mask out all the reactive tokens, and feed the active and reactive tokens to ACF and MUM. Second, we take out the tokens that need to be unmasked from the predefined permuted sequences, and send them into the diffusion as conditions. Next, the tokens after sampling are regarded as the new reactive tokens, and the other tokens are masked again until the completion of the $T$ iterations.
As for the diffusion procedure, we sample ${x}_{n}^{T}$ from a random Gaussian noise $\mathcal{N}(\mathbf{0}, \mathbf{I})$.
This initial sample is then progressively denoised in a step-by-step manner, transforming ${x}_{n}^{T}$ to ${x}_{n}^{0}$ through the iterative process defined by
\begin{equation}
\resizebox{0.85\linewidth}{!}{$
\begin{aligned}
\label{eq:infer}
{x}_{n}^{t-1}=\frac{1}{\sqrt{\alpha_{t}}}\left({x}_{n}^{t}-\frac{1-\alpha_{t}}{\sqrt{1-\bar{\alpha}_{t}}} \epsilon_{\theta}\left({x}_{n}^{t} | t, z_{n}\right)\right)+\sigma_{t} \epsilon, 
\end{aligned}
$}
\end{equation}
where $\sigma_{t}$ represents the noise level at time step $t$, and $\epsilon$ is sampled from the Gaussian distribution $\mathcal{N}(\mathbf{0}, \mathbf{I})$. 
Ultimately, the decoder in UD-VAE is used to decode all tokens and convert them back into reaction sequences. 

%% file: sec/4_experiments.tex
\section{Experiments}

\subsection{Experimental Setup}

\noindent\textbf{Datasets.} 
Following~\cite{xu2024regennet}, we evaluate our framework on NTU120-AS~\cite{xu2024regennet}, Chi3D-AS~\cite{xu2024regennet} and Inter-X~\cite{xu2024interx} using SMPL-X~\cite{SMPL-X-2019} models and actor-reactor annotations. We don't use the InterHuman dataset because it does not contain hand motion information, making it unsuitable for evaluating hand contact between two individuals. NTU120-AS contains 8,118 interaction sequences from 26 actions, captured by 3 cameras. Following~\cite{xu2024regennet}, we use camera 1 and the cross-subject protocol (half subjects for training, half for testing). Chi3D-AS contains 373 interaction sequences, which are randomly divided into training and test sets at a ratio of 4:1. \re{Inter-X includes $\sim$11K interaction sequences and more than 8.1M frames.} In all experiments, the 6D rotation representation is utilized.

\noindent\textbf{Metrics.} 
We employ the same evaluation metrics as~\cite{xu2024regennet}, which are as follows: (1) \emph{Frechet Inception Distance (FID)}: measures the latent distribution distance between the generated dataset and the real dataset. (2) \emph{Accuracy of action recognition (Acc)}: assesses action-motion matching. (3) \emph{Diversity (Div)}: measures motion diversity in the generated motion dataset. (4) \emph{Multimodality (Multimod)}: indicates diversity within the same active motion.
We generate 1,000 samples 20 times with different random seeds and report the average along with the 95\% confidence interval.

\noindent\textbf{Settings.} 
The \emph{online} setting refers to instant human action-reaction synthesis without the ability to see future motions. Conversely, the \emph{offline} setting alleviates the need for strict synchronicity. In the \emph{train-conditioned} setting, the actor motions are sampled from the training set. As for the \emph{test-conditioned} setting, the actor motions are sampled from the testing set. Following~\cite{xu2024regennet}, we mainly focus on the challenging \emph{online} scenario due to its practicality for real-world applications.

\subsection{Implementation Details.}
For the first stage, the tokenizer consists of a 3-layer encoder-decoder architecture, featuring a hidden dimension of 256, and the batch size is set to 256.
For the reaction generation, the transformer consists of 8 layers with a latent dimension of 384, and the diffusion MLPs are composed of 3 layers, each with a hidden dimension of 1024. The AdamW optimizer~\cite{AdamW-2017} is utilized with $\beta_1=0.5$ and $\beta_2=0.99$ and the batch size for this configuration is set to 128.
For the training of all models, the learning rate attains 2e-4 following 1000 iterations, utilizing a linear warm-up schedule.

\subsection{Comparisons with State-of-the-arts}
\subsubsection{\textbf{Quantitative Results on NTU120-AS}} 
The overall results are shown in~\cref{tab:cmdm}.
In the train-conditioned setting, our method achieves state-of-the-art performance in terms of FID, Acc, and Multimod on the NTU120-AS dataset. This demonstrates that our model effectively learns the reaction patterns.
As for the test-conditioned setting, our method achieves the best FID, Acc, and Multimod, especially FID, which is much better than other VAE-based and diffusion-based methods, which verifies that our method has better generalization ability. 

\begin{table*}[htbp]
  \caption[caption]{\textbf{Comparison in the \textit{online} setting on \textbf{NTU120-AS~\cite{xu2024regennet}}} for human action–reaction synthesis. $\pm$ indicates 95\% confidence interval, $\rightarrow$ means that closer to Real is better. \textbf{Bold} indicates best result and \underline{underline} indicates second best.
  }
  \label{tab:cmdm}
  \begin{center}
  \resizebox{1\textwidth}{!}{
  \begin{tabular}{l c c c c c c c c }
  \toprule
  \multirow{2}{*}{Method} & \multicolumn{4}{c}{Train conditioned} & \multicolumn{4}{c}{Test conditioned} \\
  \cmidrule(rl){2-5} \cmidrule(rl){6-9}
  & FID$\downarrow$ & Acc.$\uparrow$ & Div.$\rightarrow$ & Multimod.$\rightarrow$ & FID$\downarrow$ & Acc.$\uparrow$ & Div.$\rightarrow$ & Multimod.$\rightarrow$ \\
  \midrule
  Real & $0.09^{\pm0.00}$ & $1.000^{\pm0.0000}$ & $10.54^{\pm0.06}$ & $26.71^{\pm0.62}$ & $0.09^{\pm0.00}$ & $0.867^{\pm0.0002}$ & $13.06^{\pm0.09}$ & $25.03^{\pm0.23}$ \\
  \midrule
  cVAE~\cite{kingma2013cvae} & $77.52^{\pm7.25}$ & $0.899^{\pm0.0002}$ & $10.10^{\pm0.02}$ & $19.38^{\pm0.16}$ & $70.10^{\pm3.42}$ & $0.724^{\pm0.0002}$ & $11.14^{\pm0.04}$ & $18.42^{\pm0.26}$\\
  AGRoL~\cite{du2023agrol} & $38.04^{\pm1.45}$ & $0.932^{\pm0.0001}$ & $10.95^{\pm0.07}$ & $21.44^{\pm0.34}$ & $44.94^{\pm2.46}$ & $0.680^{\pm0.0001}$ & $\underline{12.51^{\pm0.09}}$ & $19.73^{\pm0.17}$ \\
  MDM~\cite{tevet2209mdm} & $40.13^{\pm3.65}$ & $0.955^{\pm0.0001}$ & $\mathbf{10.53^{\pm0.04}}$ & $21.15^{\pm0.26}$ & $54.54^{\pm3.94}$ & $0.704^{\pm0.0003}$ & $11.98^{\pm0.07}$ & $19.45^{\pm0.20}$ \\
  MDM-GRU~\cite{tevet2209mdm} & $5.31^{\pm0.18}$ & $\underline{0.993^{\pm0.0000}}$ & $11.03^{\pm0.06}$ & $25.04^{\pm0.36}$ & $24.25^{\pm1.39}$ & $0.720^{\pm0.0002}$ & $\mathbf{13.43^{\pm0.09}}$ & $22.24^{\pm0.29}$\\
  ReGenNet~\cite{xu2024regennet} & {$\underline{0.90^{\pm0.01}}$} & {$\mathbf{1.000^{\pm0.0000}}$} & $10.69^{\pm0.05}$ & $\underline{26.25^{\pm0.35}}$ & $\underline{11.00^{\pm0.74}}$ & $\underline{0.749^{\pm0.0002}}$ & $13.80^{\pm0.16}$ & $\underline{22.90^{\pm0.14}}$ \\
  \cellcolor{Gray}MARRS (Ours) & \cellcolor{Gray}$\mathbf{0.36^{\pm0.02}}$ & \cellcolor{Gray}$\mathbf{1.000^{\pm0.0000}}$ & \cellcolor{Gray}$\underline{10.57^{\pm0.06}}$ & \cellcolor{Gray}$\mathbf{26.53^{\pm0.20}}$ & \cellcolor{Gray}$\mathbf{9.31^{\pm0.36}}$ & \cellcolor{Gray}$\mathbf{0.753^{\pm0.0003}}$ & \cellcolor{Gray}$14.05^{\pm0.09}$ & \cellcolor{Gray}$\mathbf{23.78^{\pm0.31}}$ \\
  \bottomrule
  \end{tabular}
  }
  \end{center}
\end{table*}

\subsubsection{\textbf{Quantitative Results on Chi3D-AS}} 
To evaluate the generalizability of our approach, MARRS, we perform corresponding experiments on another dataset, Chi3D-AS~\cite{xu2024regennet}. The results are shown in~\cref{tab:chi3d}. The open-source model of ReGenNet performs poorly on Chi3D-AS, and its results are different from those reported in the paper. However, the authors did not address this issue. Therefore, we re-train ReGenNet and the corresponding evaluation model using its open source code. We maintain the same experimental setup as described in the paper. Our proposed MARRS outperforms both MDM~\cite{tevet2209mdm} and ReGenNet~\cite{xu2024regennet} across various metrics.
\begin{table*}[htbp]
  \caption[caption]{\textbf{Comparison in the \textit{online} setting on \textbf{Chi3D-AS~\cite{xu2024regennet}}} for human action-reaction synthesis. $\pm$ indicates 95\% confidence interval, $\rightarrow$ means that closer to Real is better. \textbf{Bold} indicates best result and \underline{underline} indicates second best.
  }
  \label{tab:chi3d}
  \begin{center}
  \resizebox{1.0\textwidth}{!}{
  \begin{tabular}{l c c c c c c c c }
  \toprule
  \multirow{2}{*}{Method} & \multicolumn{4}{c}{Train conditioned} & \multicolumn{4}{c}{Test conditioned} \\
  \cmidrule(rl){2-5} \cmidrule(rl){6-9}
  & FID$\downarrow$ & Acc.$\uparrow$ & Div.$\rightarrow$ & Multimod.$\rightarrow$ & FID$\downarrow$ & Acc.$\uparrow$ & Div.$\rightarrow$ & Multimod.$\rightarrow$ \\
  \midrule
  Real & $0.19^{\pm0.01}$ & $1.000^{\pm0.0000}$ & $6.59^{\pm0.21}$ & $20.96^{\pm0.46}$ & $0.98^{\pm0.22}$ & $0.602^{\pm0.0070}$ & $7.71^{\pm0.29}$ & $12.66^{\pm0.35}$ \\
  \midrule
  MDM~\cite{tevet2209mdm} & $12.64^{\pm1.85}$ & $\mathbf{1.000^{\pm0.0000}}$ & $6.88^{\pm0.24}$ & $19.06^{\pm0.29}$ & $30.54^{\pm3.49}$ & $\underline{0.443^{\pm0.0072}}$ & $6.51^{\pm0.59}$ & $10.23^{\pm0.65}$ \\
  ReGenNet~\cite{xu2024regennet} & {$\underline{0.28^{\pm0.01}}$} & {$\mathbf{1.000^{\pm0.0000}}$} & $\underline{6.70^{\pm0.13}}$ & $\underline{20.75^{\pm0.59}}$ & $\underline{21.24^{\pm3.42}}$ & $0.442^{\pm0.0053}$ & $\underline{6.96^{\pm0.65}}$ & $\underline{10.61^{\pm0.90}}$ \\
  \cellcolor{Gray}MARRS (Ours) & \cellcolor{Gray}$\mathbf{0.21^{\pm0.01}}$ & \cellcolor{Gray}$\mathbf{1.000^{\pm0.0000}}$ & \cellcolor{Gray}$\mathbf{6.69^{\pm0.14}}$ & \cellcolor{Gray}$\mathbf{20.90^{\pm0.57}}$ & \cellcolor{Gray}$\mathbf{18.94^{\pm2.21}}$ & \cellcolor{Gray}$\mathbf{0.480^{\pm0.0061}}$ & \cellcolor{Gray}$\mathbf{7.22^{\pm0.53}}$& \cellcolor{Gray}$\mathbf{11.13^{\pm0.66}}$ \\
  \bottomrule
  \end{tabular}
  }
  \end{center}
\end{table*}

\subsubsection{\textbf{\reTwo{Quantitative Results on Inter-X}}} 
\reTwo{We further conduct experiments on the Inter-X dataset and compare our method with ReGenNet~\cite{xu2024regennet} and InterMask~\cite{javed2025intermask} to provide a more comprehensive evaluation of our model. As shown in~\cref{tab:exp_Inter_X}, our method achieves superior performance, especially in the FID metric.}
\begin{table*}[htbp]
  \caption[caption]{\reTwo{\textbf{Comparison in the \textit{online} setting on Inter-X~\cite{xu2024interx}} for human action-reaction synthesis. $\pm$ indicates 95\% confidence interval, $\rightarrow$ means that closer to Real is better. \textbf{Bold} indicates best result and \underline{underline} indicates second best.}}
  \label{tab:exp_Inter_X}  
  \begin{center}
  \resizebox{1.0\textwidth}{!}{
  \begin{tabular}{l c c c c c c c c }
  \toprule
  \multirow{2}{*}{Method} & \multicolumn{4}{c}{Train conditioned} & \multicolumn{4}{c}{Test conditioned} \\
  \cmidrule(rl){2-5} \cmidrule(rl){6-9}
  & FID$\downarrow$ & Acc.$\uparrow$ & Div.$\rightarrow$ & Multimod.$\rightarrow$ & FID$\downarrow$ & Acc.$\uparrow$ & Div.$\rightarrow$ & Multimod.$\rightarrow$ \\
  \midrule
  Real & $0.13^{\pm0.01}$ & $0.998^{\pm0.0000}$ & $13.46^{\pm0.01}$ & $23.16^{\pm0.24}$ & $0.29^{\pm0.00}$ & $0.827^{\pm0.0001}$ & $15.42^{\pm0.03}$ & $22.13^{\pm0.19}$ \\
  \midrule
  ReGenNet~\cite{xu2024regennet} & $\underline{1.94^{\pm0.05}}$ & $\underline{0.974^{\pm0.0001}}$ & $\underline{13.56^{\pm0.03}}$ & $22.80^{\pm0.34}$ & 
  $\underline{10.08^{\pm0.54}}$ & $\underline{0.571^{\pm0.0004}}$ & $14.89^{\pm0.06}$ & $19.66^{\pm0.14}$ \\
  InterMask~\cite{javed2025intermask} & $2.26^{\pm0.06}$ & $0.853^{\pm0.0001}$ & $13.59^{\pm0.04}$ & $\underline{23.32^{\pm0.31}}$ & $14.66^{\pm0.56}$ & $0.503^{\pm0.0004}$ & $\underline{14.92^{\pm0.09}}$ & $\underline{20.01^{\pm0.40}}$ \\
  \cellcolor{Gray}MARRS (Ours) & \cellcolor{Gray}$\mathbf{0.49^{\pm0.00}}$ & \cellcolor{Gray}$\mathbf{0.987^{\pm0.0000}}$ & \cellcolor{Gray}$\mathbf{13.48^{\pm0.04}}$ & \cellcolor{Gray}$\mathbf{23.28^{\pm0.21}}$ & 
  \cellcolor{Gray}$\mathbf{6.15^{\pm0.46}}$ & \cellcolor{Gray}$\mathbf{0.636^{\pm0.0002}}$ & \cellcolor{Gray}$\mathbf{15.07^{\pm0.04}}$& \cellcolor{Gray}$\mathbf{20.66^{\pm0.33}}$ \\
  \bottomrule
  \end{tabular}
  }
  \end{center}
\end{table*}

\subsubsection{\textbf{Offline Setting on NTU120-AS}}
To demonstrate the adaptability and generalization of our proposed method MARRS, we additionally provide the offline setting results on NTU120-AS in~\cref{tab:cmdm_offline}. For the model, we simply adjust the attention mask in the transformer to fit the offline setting. As shown in the table, MARRS achieves optimal performance on FID, Acc and competitive results on Div and Multimod, highlighting the efficacy of our approach and its adaptability.
\begin{table}[htbp]
  \caption[caption]{\textbf{Comparison in the offline setting on NTU120-AS~\cite{xu2024regennet}.} $\pm$ indicates 95\% confidence interval, $\rightarrow$ means that closer to Real is better. \textbf{Bold} indicates the best and \underline{underline} indicates second best.}
  \label{tab:cmdm_offline}
  
  \begin{center}
  \resizebox{0.49\textwidth}{!}{
  \begin{tabular}{l c c c c}
  \toprule
  Method & FID$\downarrow$ & Acc.$\uparrow$ & Div.$\rightarrow$ & Multimod.$\rightarrow$ \\
  \midrule
  Real & $0.09^{\pm0.00}$ & $0.867^{\pm0.0002}$ & $13.06^{\pm0.09}$ & $25.03^{\pm0.23}$ \\
  \midrule
  cVAE~\cite{kingma2013cvae} &$74.73^{\pm4.86}$ & $0.760^{\pm0.0002}$ & $11.14^{\pm0.04}$ & $18.40^{\pm0.26}$ \\
  AGRoL~\cite{du2023agrol} & $16.55^{\pm1.41}$ & $0.716^{\pm0.0002}$ & $13.84^{\pm0.10}$ & $21.73^{\pm0.20}$ \\
  MDM~\cite{tevet2209mdm} & $7.49^{\pm0.62}$ & $\underline{0.775^{\pm0.0003}}$ & $\underline{13.67^{\pm0.18}}$ & $24.14^{\pm0.29}$ \\
  MDM-GRU~\cite{tevet2209mdm} & $24.25^{\pm1.39}$ & $0.720^{\pm0.0002}$ & $\mathbf{13.43^{\pm0.09}}$ & $22.24^{\pm0.29}$ \\
  ReGenNet~\cite{xu2024regennet} & \underline{$6.19^{\pm0.33}$} & $0.772^{\pm0.0003}$ & $14.03^{\pm0.09}$ & $\mathbf{25.21^{\pm0.34}}$ \\
  \cellcolor{Gray}MARRS (Ours) & \cellcolor{Gray}$\mathbf{5.93^{\pm0.18}}$ & \cellcolor{Gray}$\mathbf{0.783^{\pm0.0002}}$ & \cellcolor{Gray}$13.96^{\pm0.17}$ & \cellcolor{Gray}$\underline{25.38^{\pm0.50}}$ \\
  \bottomrule
  \end{tabular}
  }
  \end{center}
\end{table}

\subsubsection{\textbf{Qualitative Comparisons}} 
\begin{figure*}[htbp]
  \centering
   \includegraphics[width=0.9\linewidth,page=1]{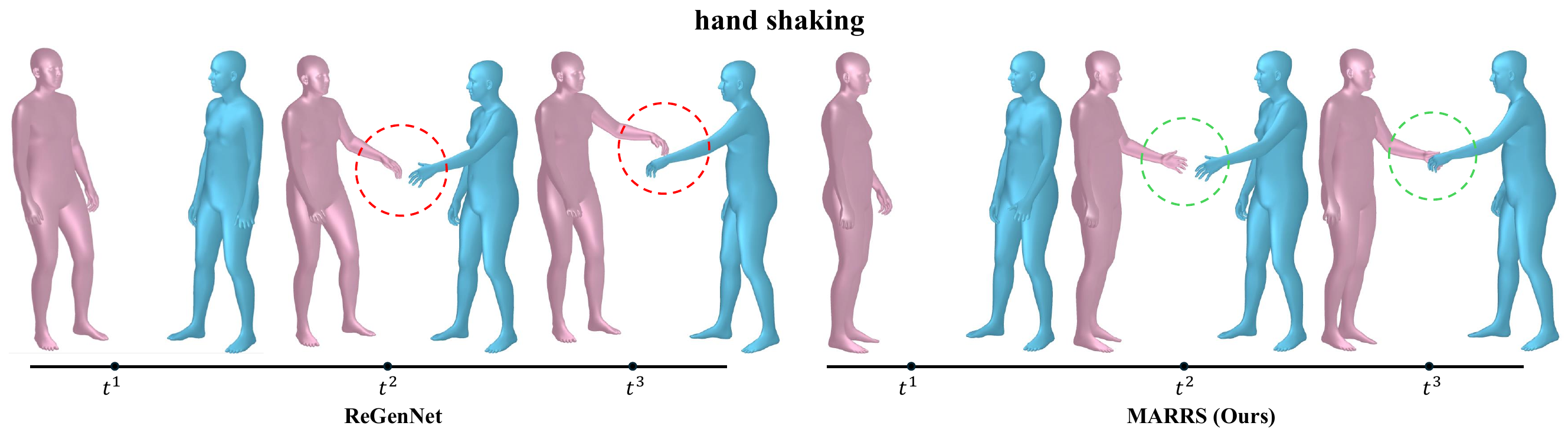}
  \includegraphics[width=0.9\linewidth,page=2]{figs/cmp_regen_mar_v3_crop.pdf}
  \includegraphics[width=0.9\linewidth,page=3]{figs/cmp_regen_mar_v3_crop.pdf}
   \caption{\re{\textbf{Visualization Comparison with RegenNet on NTU120-AS}}. {\color[RGB]{79,153,178}\textbf{Blue}} \re{for actors and} {\color[RGB]{185,142,162}\textbf{Red}} \re{for reactors.} \re{Our method produces more plausible body movements and relative positions, as well as more natural hand gestures of reactors. The} {\color[RGB]{255,0,0}red} \re{dashed boxes highlight artifacts, while the} {\color[RGB]{71,212,90}green} \re{dashed boxes indicate more reasonable results.}}
   \label{fig:Vis_CMP_ReGenNet}
\end{figure*}

\begin{figure*}[htbp]
  \centering
   \includegraphics[width=0.9\linewidth,page=1]{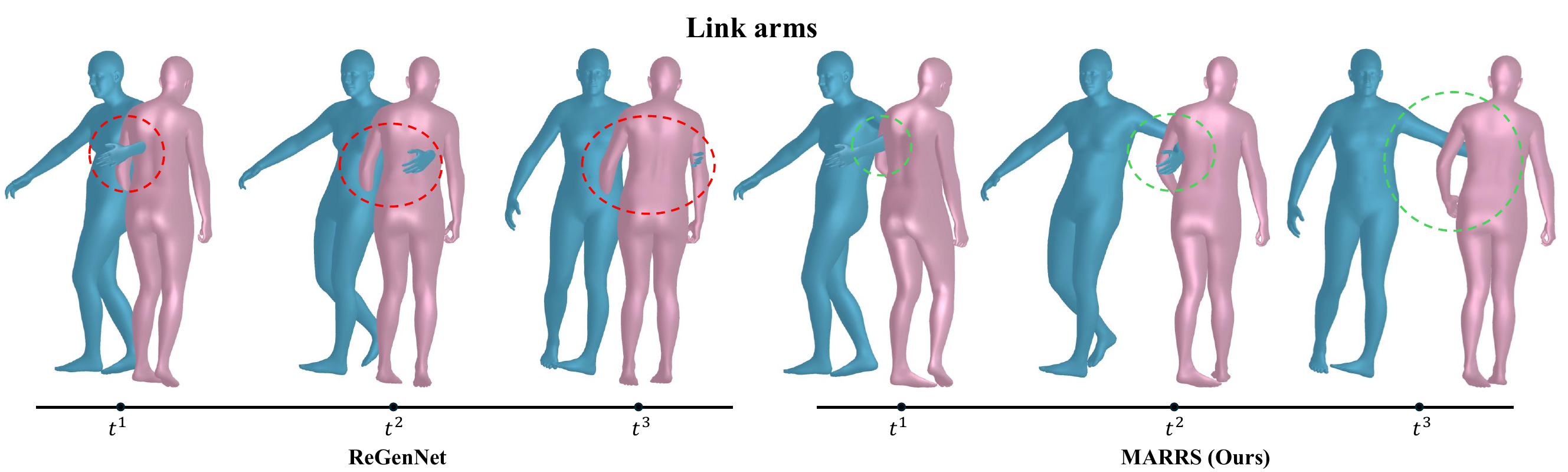}
  \includegraphics[width=0.9\linewidth,page=2]{figs/InterX_cmp_regen_mar_v1_crop.pdf}
   \caption{\re{\textbf{Visualization Comparison with RegenNet on Inter-X}}. {\color[RGB]{79,153,178}\textbf{Blue}} \re{for actors and} {\color[RGB]{185,142,162}\textbf{Red}} \re{for reactors.} \re{Our method produces more plausible body movements and relative positions, as well as more natural hand gestures of reactors. The} {\color[RGB]{255,0,0}red} \re{dashed boxes highlight artifacts, while the} {\color[RGB]{71,212,90}green} \re{dashed boxes indicate more reasonable results.}}
   \label{fig:Vis_CMP_ReGenNet_InterX}
\end{figure*}

In \cref{fig:Vis_CMP_ReGenNet} and \cref{fig:Vis_CMP_ReGenNet_InterX}, we provide a qualitative comparison between ReGenNet~\cite{xu2024regennet} and MARRS on NTU120-AS and Inter-X, respectively. It can be seen that the human reaction synthesized by our method has more reasonable relative positions and body movements. In particular, our method can generate more natural and plausible hand gestures by modeling the body and hands effectively. We provide multiple \textbf{demo} videos in the supplementary materials.

\begin{table*}[htbp]
    \caption[caption]{\re{\textbf{Ablation studies on the Unit Division.} ``w.o. Unit Division" represents keeping whole-body as a complete unit. ``Upper \& Lower" represents dividing whole-body into the upper-and-lower-body proposed by SCA~\cite{ghosh2021sca}. ``6-Unit Division" represents dividing whole-body into six separate units proposed by ParCo~\cite{zou2024parco}. Our proposed approach achieves a more balanced performance overall under train-conditioned and test-conditioned settings.}} 
  \label{tab:ablations_splitpart}
  \begin{center}
  \resizebox{1\textwidth}{!}{
  \begin{tabular}{l c c c c c c c c c}
  \toprule
   \multirow{2}{*}{Settings} & \multirow{2}{*}{\makecell{Parameters\\(M)}} & \multicolumn{4}{c}{Train conditioned} & \multicolumn{4}{c}{Test conditioned} \\
   \cmidrule(rl){3-6} \cmidrule(rl){7-10}
  & & FID$\downarrow$ & Acc.$\uparrow$ & Div.$\rightarrow$ & Multimod.$\rightarrow$ & FID$\downarrow$ & Acc.$\uparrow$ & Div.$\rightarrow$ & Multimod.$\rightarrow$ \\
  \cmidrule{1-2} \cmidrule{3-6} \cmidrule{7-10}
  Real & - & $0.09^{\pm0.00}$ & $1.000^{\pm0.0000}$ & $10.54^{\pm0.06}$ & $26.71^{\pm0.62}$ & $0.09^{\pm0.00}$ & $0.867^{\pm0.0002}$ & $13.06^{\pm0.09}$ & $25.03^{\pm0.23}$ \\
  \midrule
  w.o. Unit Division & 85.7 &$0.79^{\pm0.01}$ & $\mathbf{1.000^{\pm0.0000}}$ & $10.62^{\pm0.02}$ & $26.25^{\pm0.26}$ & $15.00^{\pm0.80}$ & $0.745^{\pm0.0002}$ & $\underline{13.87^{\pm0.15}}$ & $23.52^{\pm0.21}$\\
  Upper \& Lower~\cite{ghosh2021sca} & 90.2 &$0.72^{\pm0.25}$ & $\mathbf{1.000^{\pm0.0000}}$ & $\underline{10.58^{\pm0.05}}$ & $26.27^{\pm0.31}$ & $\underline{10.21^{\pm0.26}}$ & $0.745^{\pm0.0002}$ & $14.08^{\pm0.19}$ & $\mathbf{23.79^{\pm0.24}}$ \\
  6-Unit Division~\cite{zou2024parco} & 207.0 &$\mathbf{0.21^{\pm0.00}}$ & $\mathbf{1.000^{\pm0.0000}}$ & $10.50^{\pm0.04}$ & $\underline{26.52^{\pm0.31}}$ & $14.97^{\pm0.82}$ & $\mathbf{0.769^{\pm0.0001}}$ & $\mathbf{13.30^{\pm0.13}}$ & $23.01^{\pm0.30}$ \\
   \cellcolor{Gray}Body \& Hands (Ours) & 
   \cellcolor{Gray}90.2 & \cellcolor{Gray}$\underline{0.36^{\pm0.02}}$ & \cellcolor{Gray}$\mathbf{1.000^{\pm0.0000}}$ & \cellcolor{Gray}$\mathbf{10.57^{\pm0.06}}$ & \cellcolor{Gray}$\mathbf{26.53^{\pm0.20}}$ & \cellcolor{Gray}$\mathbf{9.31^{\pm0.36}}$ & \cellcolor{Gray}$\underline{0.753^{\pm0.0003}}$ & \cellcolor{Gray}$14.05^{\pm0.09}$ & \cellcolor{Gray}$\underline{23.78^{\pm0.31}}$ \\
  \bottomrule
  \end{tabular}
  }
  \end{center}
\end{table*}

\begin{table*}[htbp]
  \caption[caption]{\re{\textbf{Ablation studies on Action-Conditioned Fusion (ACF)}. ``Concatenate Fuse" is a method proposed by ReGenNet~\cite{xu2024regennet} for a reactor to get information from an actor. ``Cooperative Transformer" is a transformer structure designed by InterGen~\cite{liang2024intergen}. Our approach achieves SoTA on almost all metrics.}}
  \label{tab:ablations_interaction}
  \begin{center}
  \resizebox{1.0\textwidth}{!}{
  \begin{tabular}{l c c c c c c c c}
  \toprule
   \multirow{2}{*}{Settings} & \multicolumn{4}{c}{Train conditioned} & \multicolumn{4}{c}{Test conditioned} \\
  \cmidrule(rl){2-5} \cmidrule(rl){6-9}
  & FID$\downarrow$ & Acc.$\uparrow$ & Div.$\rightarrow$ & Multimod.$\rightarrow$ & FID$\downarrow$ & Acc.$\uparrow$ & Div.$\rightarrow$ & Multimod.$\rightarrow$ \\
  \midrule
  Real & $0.09^{\pm0.00}$ & $1.000^{\pm0.0000}$ & $10.54^{\pm0.06}$ & $26.71^{\pm0.62}$ & $0.09^{\pm0.00}$ & $0.867^{\pm0.0002}$ & $13.06^{\pm0.09}$ & $25.03^{\pm0.23}$ \\
  \midrule
   Concatenate Fuse~\cite{xu2024regennet} &$\underline{0.54^{\pm0.01}}$  & $\underline{0.999^{\pm0.0001}}$ & $10.64^{\pm0.06}$ & $\underline{26.47^{\pm0.20}}$ & $11.30^{\pm0.70}$ & $0.743^{\pm0.0002}$ & $14.10^{\pm0.10}$ & $\mathbf{24.41^{\pm0.51}}$ \\
  Cooperative
Transformer~\cite{liang2024intergen} & $0.59^{\pm0.23}$ & $\mathbf{1.000^{\pm0.0000}}$ & $\underline{10.60^{\pm0.05}}$ & $26.29^{\pm0.32}$ & $\underline{10.27^{\pm0.27}}$ & $\underline{0.752^{\pm0.0003}}$ & $\mathbf{13.85^{\pm0.17}}$ & $\underline{23.79^{\pm0.20}}$ \\
  \cellcolor{Gray}ACF (Ours) & \cellcolor{Gray}$\mathbf{0.36^{\pm0.02}}$ & \cellcolor{Gray}$\mathbf{1.000^{\pm0.0000}}$ & \cellcolor{Gray}$\mathbf{10.57^{\pm0.06}}$ & \cellcolor{Gray}$\mathbf{26.53^{\pm0.20}}$ & \cellcolor{Gray}$\mathbf{9.31^{\pm0.36}}$ & \cellcolor{Gray}$\mathbf{0.753^{\pm0.0003}}$ & \cellcolor{Gray}$\underline{14.05^{\pm0.09}}$ & \cellcolor{Gray}$23.78^{\pm0.31}$ \\
  
  \bottomrule
  \end{tabular}
  }
  \end{center}

\end{table*}

\begin{table*}[htbp]
    \caption[caption]{\textbf{Ablation studies on Mutual Unit Modulation (MUM)}. ``w.o. Unit Communication" indicates that the body and hands branches do not interact with each other. ``Coordination Layer'' is a method proposed by ParCo~\cite{zou2024parco} for transferring information between different units. ``Hands $\rightarrow$ Body'' and ``Body $\rightarrow$ Hands'' represent a unidirectional transfer of information between units. Our approach achieves SoTA on almost all metrics.}
  \label{tab:ablations_partfuse}
  
  \begin{center}
  \resizebox{1.0\textwidth}{!}{
  \begin{tabular}{l c c c c c c c c}
  \toprule
   \multirow{2}{*}{Settings} & \multicolumn{4}{c}{Train conditioned} & \multicolumn{4}{c}{Test conditioned} \\
  \cmidrule(rl){2-5} \cmidrule(rl){6-9}
  & FID$\downarrow$ & Acc.$\uparrow$ & Div.$\rightarrow$ & Multimod.$\rightarrow$ & FID$\downarrow$ & Acc.$\uparrow$ & Div.$\rightarrow$ & Multimod.$\rightarrow$ \\
  \midrule
  Real & $0.09^{\pm0.00}$ & $1.000^{\pm0.0000}$ & $10.54^{\pm0.06}$ & $26.71^{\pm0.62}$ & $0.09^{\pm0.00}$ & $0.867^{\pm0.0002}$ & $13.06^{\pm0.09}$ & $25.03^{\pm0.23}$ \\
  \midrule
  w.o. Unit Communication& $0.86^{\pm0.43}$ & $\mathbf{1.000^{\pm0.0000}}$ & $10.63^{\pm0.05}$ & $26.16^{\pm0.33}$ & $13.09^{\pm0.48}$ & $0.738^{\pm0.0002}$ & $\underline{13.90^{\pm0.12}}$ & $23.16^{\pm0.39}$ \\
   Coordination Layer~\cite{zou2024parco} & $0.61^{\pm0.01}$ & $\mathbf{1.000^{\pm0.0000}}$ & $10.62^{\pm0.04}$ & $26.30^{\pm0.32}$ & $\underline{10.06^{\pm0.46}}$ & $0.739^{\pm0.0002}$ & $14.18^{\pm0.17}$ & $\underline{23.60^{\pm0.27}}$ \\
   MUM (Hands $\rightarrow$ Body) &$0.85^{\pm0.01}$  & $\underline{0.999^{\pm0.0000}}$ & $\underline{10.61^{\pm0.05}}$ & $26.22^{\pm0.16}$ & $13.69^{\pm0.93}$ & $\underline{0.744^{\pm0.0004}}$ & $\mathbf{13.72^{\pm0.11}}$ & $22.78^{\pm0.19}$ \\
   MUM (Body $\rightarrow$ Hands) &$\underline{0.57^{\pm0.01}}$  & $\mathbf{1.000^{\pm0.0000}}$ & $10.63^{\pm0.06}$ & $\underline{26.36^{\pm0.19}}$ & $10.53^{\pm0.61}$ & $0.734^{\pm0.0003}$ & $14.20^{\pm0.12}$ & $23.30^{\pm0.28}$ \\
  \cellcolor{Gray}MUM (Ours) & \cellcolor{Gray}$\mathbf{0.36^{\pm0.02}}$ & \cellcolor{Gray}$\mathbf{1.000^{\pm0.0000}}$ & \cellcolor{Gray}$\mathbf{10.57^{\pm0.06}}$ & \cellcolor{Gray}$\mathbf{26.53^{\pm0.20}}$ & \cellcolor{Gray}$\mathbf{9.31^{\pm0.36}}$ & \cellcolor{Gray}$\mathbf{0.753^{\pm0.0003}}$ & \cellcolor{Gray}$14.05^{\pm0.09}$ & \cellcolor{Gray}$\mathbf{23.78^{\pm0.31}}$ \\
  \bottomrule
  \end{tabular}
  }
  \end{center}

\end{table*}

\subsection{Ablation Study}
In this section, we carry out extensive ablation experiments to investigate the effectiveness of the crucial components in MARRS, thus offering a more profound understanding of our method. Unless specified otherwise, we conduct all the ablations on the NTU120-AS dataset under the \emph{online} setting. 

\subsubsection{\textbf{Unit Division of Body and Hands}}
In order to explore the necessity of dividing the whole-body into the body and hands units in the human reaction synthesis, we conduct the experiments on three methods: (\emph{i}) keeping whole-body as a complete unit without unit division, (\emph{ii}) dividing it into the upper-and-lower-body proposed by SCA~\cite{ghosh2021sca}, (\emph{iii}) dividing it into six separate units proposed by ParCo~\cite{zou2024parco}. Except for three different dividing ways, the training of the first stage VAE and the structure of the second stage model are exactly the same. The results are shown in~\cref{tab:ablations_splitpart}. There is a significant decrease in FID, Acc, and MultiMod in the test-conditioned setting without unit-division. The performance of FID and MultiMod in the test-conditioned setting increase somewhat after applying the upper-and-lower-body division. When performing the 6-unit division, FID under train-conditioned and Acc under test-conditioned achieve the best results, however, FID under test-conditioned shows almost no improvement compared to w.o. unit division, and \textbf{\textit{increasing the number of division units makes training and inference more cumbersome and slower}}. In contrast, our approach achieves more balanced results under train- and test-conditioned settings, while Div and Multimod (train-conditioned) and FID (test-conditioned) achieve the best performance.

\subsubsection{\textbf{Action-Conditioned Fusion (ACF)}}
To demonstrate the effectiveness of ACF, we additionally try two actor-reactor interactions, (\emph{i}) concatenate fuse proposed by ReGenNet~\cite{xu2024regennet}, and (\emph{ii}) cooperative transformer proposed by InterGen~\cite{liang2024intergen}. The experimental results are shown in~\cref{tab:ablations_interaction}. Concatenate fuse only performs well on Multimod under test-conditioned, while the performance of all other metrics is unsatisfactory. After applying cooperative transformer, FID, Acc and Div under test-conditioned are improved to some extent. Compared with others, our proposed ACF obtains optimal performance in all four metrics under train-conditioned as well as FID and Acc under test-conditioned, proving that ACF can capture the motion patterns of the two persons well.

\subsubsection{\textbf{\reTwo{Mutual Unit Modulation (MUM)}}}
\reTwo{As noted in ParCo~\cite{zou2024parco}, inter-unit communication is crucial for stable whole-body motions. We evaluate four settings: (\emph{i}) no communication, (\emph{ii}) ParCo’s coordination layer~\cite{zou2024parco}, (\emph{iii}) MUM hands$\rightarrow$body transfer, and (\emph{iv}) MUM body$\rightarrow$hands transfer (\cref{tab:ablations_partfuse}). Without communication, performance drops under both train- and test-conditioned settings, though not catastrophically as in text-to-motion~\cite{zou2024parco}. Adding the coordination layer substantially improves FID, confirming the need for information exchange in reaction synthesis. With unidirectional MUM, hands$\rightarrow$body degrades some metrics, whereas body$\rightarrow$hands improves overall results, suggesting hand-only cues are insufficient to guide body motion while body context refines hand details.
Ultimately, the bidirectional communication of MUM achieves the best performance for all metrics under train-conditioned and FID, Acc, and Multimod under test-conditioned, demonstrating the effectiveness of our proposed MUM and the necessity of bidirectional unit communication.}

\subsubsection{\textbf{Different implementation forms of MUM}}
There are many ways to implement the functionality of MUM. The most common alternative is cross-attention. We provide a comparison of the two in the~\cref{tab:MUM_vs_cross}. \re{The motivation behind this design is as follows. ACF and MUM together form the fundamental building block of our model. Within each block, ACF is primarily responsible for modeling inter-frame information, while MUM focuses on capturing intra-frame interactions. If MUM were to employ cross-attention to explicitly model cross-frame dependencies, the learning of intra-frame interactions would become less effective. Moreover, introducing cross-attention in MUM would significantly increase the number of model parameters, making the model more prone to overfitting.}


\begin{table}[htbp]
    \caption{\textbf{Comparison of results for MUM and Cross-Attention.} MUM is lightweight and effective.
    }
    \label{tab:MUM_vs_cross}
    \centering
    \scalebox{1}{
    \begin{tabular}{c|ccc}
        \toprule
         Methods & FID$\downarrow$ &Acc.$\uparrow$ & Params (M)  \\
         \midrule
          Cross-Attention  &\et{9.89}{0.46} &\et{0.722}{.0003}   &100.6 \\
          MUM(Ours)  &\etb{9.31}{0.36} &\etb{0.753}{.0003}  & \textbf{90.2} \\
        \bottomrule
    \end{tabular}
    }

\end{table}

\subsubsection{\textbf{Diffusion for Autoregressive (AR) Reaction Generation}}\label{sec:vae-ours}
\begin{figure}[htbp]
  \centering
   \includegraphics[width=1\linewidth,page=1]{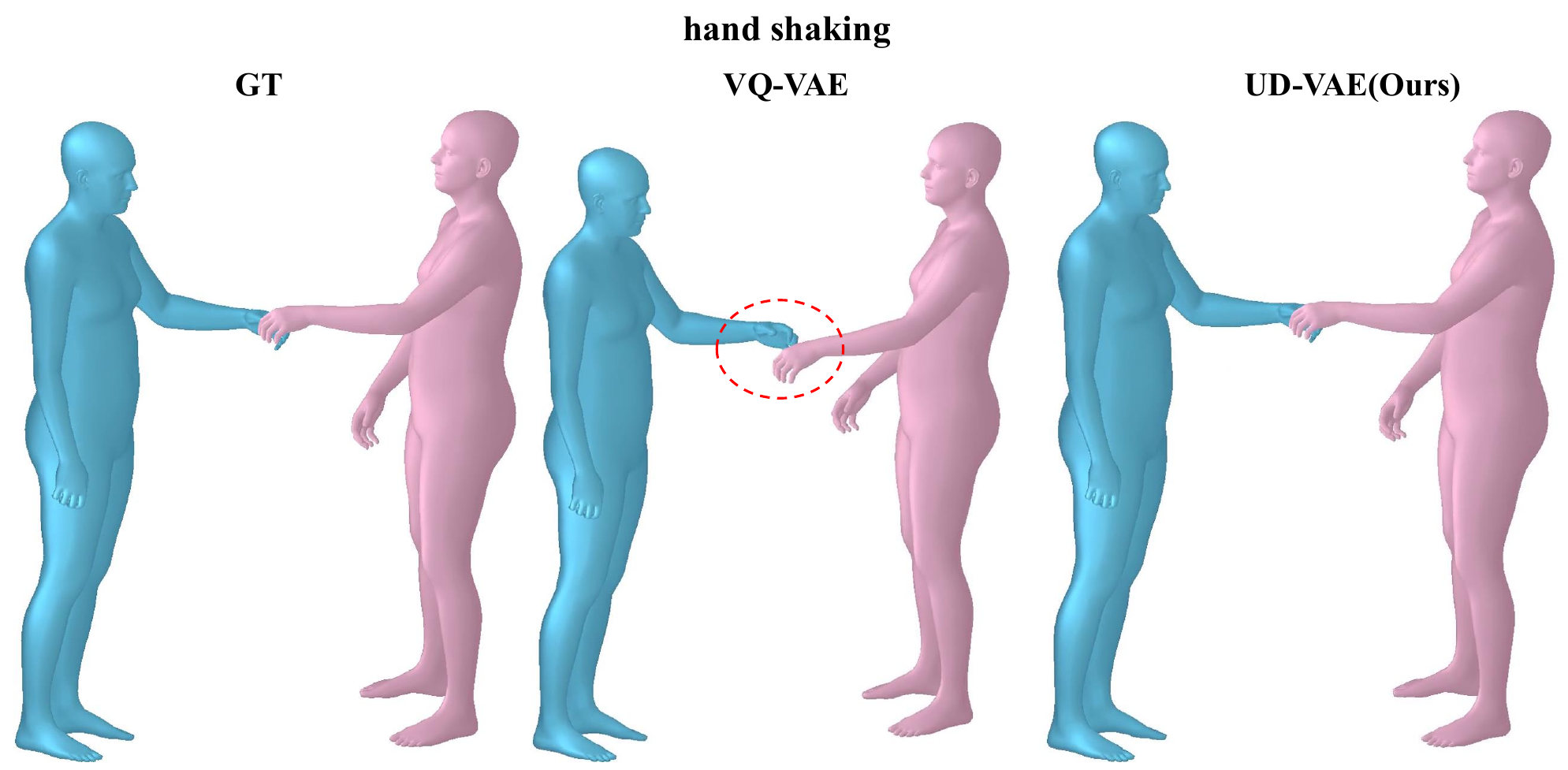}
  \includegraphics[width=1\linewidth,page=5]{figs/GT_VQVAE_UDVAE_cmp_crop.pdf}
   \caption{\re{\textbf{Visualization Comparison of Reconstruction: VQ-VAE vs. UD-VAE (Ours).} The results in the red dashed box show reconstruction artifacts by VQ-VAE, while our results align more closely with the ground truth (GT).} {\color[RGB]{79,153,178}\textbf{Blue}} \re{for actors and} {\color[RGB]{185,142,162}\textbf{Red}} \re{for reactors.} }
   \label{fig:VQVAE_UDVAE_CMP}
\end{figure}

Since AR generation is often accompanied by vector quantization (VQ), we first investigate the performance of VQ-based methods in human reaction synthesis. As shown in~\cref{tab:ablations_vq}, VQ-VAE (AR) performs poorly (10.06 FID) in the first reconstruction phase, and the failed reconstruction means that the second generation phase is not feasible. Meanwhile, we apply the unit-division way in our proposed Unit Division to VQ-VAE and name it UD-VQ-VAE (AR). The results of the reconstruction are significantly improved, but it still performs poorly in the generation phase (53.57 FID).
What is more, we apply ACTOR (VAE)~\cite{petrovich2021actor} and AMDM (AR Diffusion)~\cite{shi2024amdm} to human reaction, but neither FID nor accuracy achieves satisfactory performance.
In addition, we explore removing the diffusion loss in the second stage, but directly using L2 loss to learn continuous motion tokens, which resulted in a drastic deterioration of FID, proving that the diffusion loss is necessary. \re{A visualization comparison of reconstruction between VQ-VAE and UD-VAE is provided in~\cref{fig:VQVAE_UDVAE_CMP}}.
\begin{table}[htbp]
  \caption[caption]{\re{\textbf{Ablation studies on different framework.} ``VQ-VAE''  is based on MMM~\cite{pinyoanuntapong2024mmm}. ``UD-VQ-VAE'' is based on ``VQ-VAE'' by applying our proposed unit division (UD). ``ACTOR''~\cite{petrovich2021actor} is a VAE-based generation method. ``AMDM''~\cite{shi2024amdm} is an autoregressive (AR) diffusion method. ``L2 Loss'' means directly using L2 Loss instead of incorporating diffusion loss.}}
  \label{tab:ablations_vq}
  \begin{center}
  \resizebox{0.49\textwidth}{!}{
  \begin{tabular}{l c c c c }
  \toprule
  \multirow{2}{*}{Method} & \multicolumn{2}{c}{Reconstruction} & \multicolumn{2}{c}{Generation} \\
  \cmidrule(rl){2-5}
  & FID$\downarrow$ & Acc.$\uparrow$ & FID$\downarrow$ & Acc.$\uparrow$  \\
  \midrule
  VQ-VAE (AR) & $10.06^{\pm0.32}$ & $0.783^{\pm0.0002}$ & \makebox[1cm]{\textemdash} & \makebox[1cm]{\textemdash} \\
  UD-VQ-VAE (AR) & $5.76^{\pm0.26}$ & $0.805^{\pm0.0002}$ & $53.57^{\pm6.35}$ & $0.562^{\pm0.0004}$ \\
  ACTOR (VAE) & $\underline{1.21^{\pm0.23}}$ & $\underline{0.823^{\pm0.0002}}$ & $20.16^{\pm0.65}$ & $0.730^{\pm0.0004}$ \\
  AMDM (Diffusion) &  \makebox[1cm]{\textemdash} &  \makebox[1cm]{\textemdash} & \et{23.46}{1.39} &\et{0.718}{.0002} \\
  L2 Loss (VAE) &  $\mathbf{0.18^{\pm0.10}}$ & $\mathbf{0.866^{\pm0.0001}}$ & $\underline{15.95^{\pm1.93}}$ & $\mathbf{0.760^{\pm0.0002}}$ \\
  \cellcolor{Gray}MARRS (Ours) & \cellcolor{Gray}{$\mathbf{0.18^{\pm0.10}}$} & \cellcolor{Gray}{$\mathbf{0.866^{\pm0.0001}}$} &  \cellcolor{Gray}$\mathbf{9.31^{\pm0.36}}$ & \cellcolor{Gray}$\underline{0.753^{\pm0.0003}}$ \\
  \bottomrule
  \end{tabular}
  }
  \end{center}

\end{table}

\subsection{Accuracy of Hand Poses and Global Translation}
We used coordinate-based metrics (\textit{APE and AVE})~\cite{APE_AVE} to measure the accuracy of hand poses and global translation (root) with SOTA. As shown in~\cref{tab:hand}, our method is more accurate in hand modeling, demonstrating the effectiveness of unit division and interaction. The error of global translation is also smaller than that of ReGenNet, further highlighting the effectiveness of the MARRS framework.
\begin{table}[!h]
    \caption{ 
     \textbf{Comparison on hands and global translation.}
    }
    \label{tab:hand}
    \centering
    \scalebox{0.82}{
    \begin{tabular}{c|cccc}
        \toprule
         Methods & APE (Hands)$\downarrow$&  AVE (Hands) $\downarrow$& APE (Root)$\downarrow$&  AVE (Root) $\downarrow$   \\

         \midrule
          ReGenNet  &\et{0.420}{0.0001} &\et{0.446}{0.0355}  &\et{0.327}{0.0001} &\et{0.055}{0.0011} \\
          MARRS(Ours)  &\etb{0.388}{.0001} &\etb{0.430}{0.0357}  &\etb{0.294}{0.0001} &\etb{0.043}{0.0010} \\
        \bottomrule
    \end{tabular}
    }

\end{table}

\subsection{Model Scalability and Complexity}
\reTwo{To evaluate MARRS scalability, we present three model versions with ~30M, 60M, and 90M parameters. \cref{tab:computation} reports performance, parameters, training and inference time. We observe that larger models improve generation, particularly FID. We also compare MARRS parameters and training/inference time with ReGenNet~\cite{xu2024regennet}. Results indicate MARRS converges faster than ReGenNet, and MARRS-Tiny enables faster inference while maintaining competitive performance.}


\begin{table}[h]
    \caption{\textbf{Model scaling results and comparison of computational complexity.} Our proposed MARRS can converge faster during training than ReGenNet~\cite{xu2024regennet} and enlarging the model size can enhance the overall generation performance. \reTwo{All timing measurements are obtained using a single H20.}
    }
    \label{tab:computation}
    
    \centering
    \scalebox{0.83}{
    \begin{tabular}{c|ccccc}
        \toprule
         \multirow{2}{*}{Methods}  & \multirow{2}{*}{FID $\downarrow$}&  \multirow{2}{*}{Acc $\uparrow$}  & Params &Training &Inference \\

           ~ & & ~ &(M)  &Time (h) & Time (s) \\
       
         \midrule
          ReGenNet~\cite{xu2024regennet} & $11.00^{\pm0.74}$ & $0.749^{\pm0.0002}$ & \textbf{26.8} & 121.1 &0.058 \\
          MARRS-Tiny  &$10.55^{\pm0.62}$ & $0.743^{\pm0.0003}$ &30.3 & \textbf{10.6}& \textbf{0.039}  \\
          MARRS-Small  &$10.12^{\pm0.352}$ & $0.751^{\pm0.0003}$ &56.8 & 12.7& 0.086 \\          
          MARRS-Base  &$\mathbf{9.31^{\pm0.36}}$ & $\mathbf{0.753^{\pm0.0003}}$ &90.2 & 14.0&0.204  \\
        \bottomrule
        \end{tabular}
    }

\end{table}

\subsection{User Study}
\begin{figure}[h]
  \centering
   \includegraphics[width=1\linewidth]{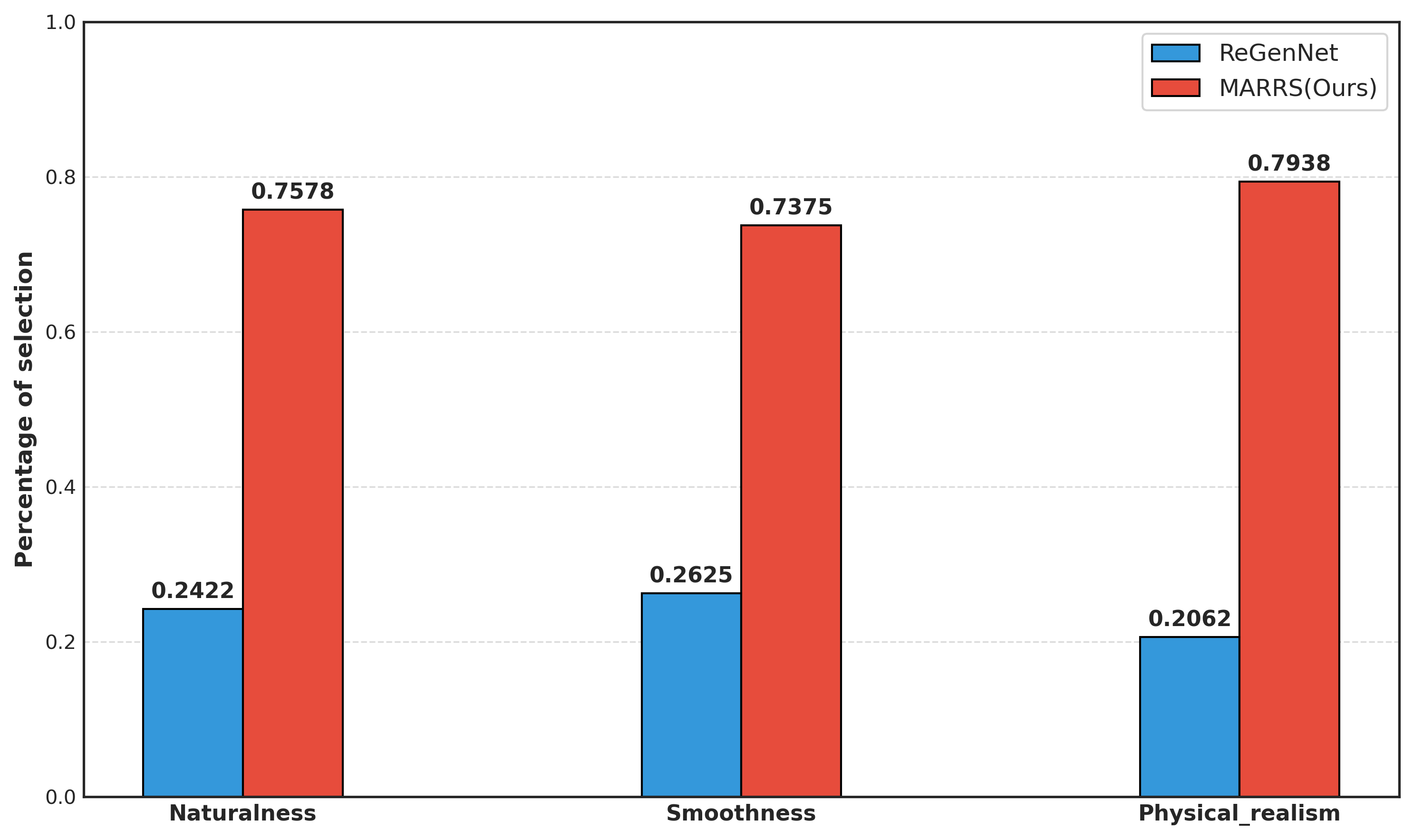}
   \caption{\textbf{User study.} We use three subjective indicators, Naturalness, Smoothness, and Realism, to compare with ReGenNet.
}
   \label{fig:user_study}
\end{figure}

We conduct a user study to evaluate MARRS against ReGenNet~\cite{xu2024regennet}. 20 users were asked to evaluate 32 samples. The results are shown in Fig.~\ref{fig:user_study}. Compared to the SOTA method, ReGenNet, about \textbf{76\%} of users believe that our motions are more \textit{natural}, about \textbf{74\%} of users believe that our motions are \textit{smoother}, and about \textbf{79\%} of users believe that our motions are more physically \textit{realistic}.

\section{Conclusion and Limitation}
\subsection{Conclusion}
In this paper, we introduce an innovative framework named MARRS, designed to generate synchronized and fine reactions. Initially, we present the UD-VAE, which divides the whole body into distinct units: body and hands, allowing for independent encoding. Following this, we introduce ACF, a process that involves randomly masking reactive tokens and then extracting information of the body and hands from the remaining active tokens. To enable whole-body collaboration, we propose MUM, which uses information from either the body or hands to adaptively adjust the other unit. Finally, in our diffusion model, we utilize a compact MLP as a noise predictor for each specific unit and integrate diffusion loss to capture the probability distribution of each token. Both quantitative and qualitative evaluations indicate that our method outperforms existing methods. 

\subsection{Limitation}
Due to the lack of high-quality human reaction datasets, we validated the method only on NTU120-AS and Chi3D-AS. Currently, methods in this field generally suffer from a certain degree of foot sliding, although our approach alleviates this issue to some extent. Further exploration of physical constraints is needed to address this problem more effectively in the future. Owing to the limited accuracy of the current dataset, the generation of certain fine-grained motions (such as finger contacts) is not sufficiently precise. Related researchers in the community are encouraged to explore more challenging aspects of the human reaction synthesis. We hope that MARRS could provide some new insight for the community.